\documentclass[11pt]{article}
\usepackage{hyperref}
\usepackage{booktabs}

\usepackage{algorithm}
\usepackage{algpseudocode}
\usepackage{float}

\usepackage{tikz}
\usetikzlibrary{positioning, arrows.meta, shapes.geometric, shapes.multipart, fit}

\usepackage{tcolorbox}
\tcbuselibrary{listings, breakable, skins}

\usepackage{amsmath, amssymb, amsthm} 
\usepackage{graphicx} 
\usepackage{geometry} 
\usepackage{cite} 
\usepackage{setspace} 
\usepackage{authblk} 

\usepackage{subcaption}
\title{Adaptive Insurance Reserving via CVaR-Constrained Reinforcement Learning under Macroeconomic Regimes}

\author{
Stella C. Dong \\
Reinsurance Analytics \\
\texttt{stella.dong@reinsuranceanalytics.io}
}

\date{}

\begin{document}

\maketitle

\begin{abstract}
We develop a reinforcement learning (RL) framework for insurance loss reserving that formulates reserve setting as a finite-horizon sequential decision problem under claim development uncertainty, macroeconomic stress, and solvency governance. The reserving process is modeled as a Markov Decision Process (MDP) in which reserve adjustments influence future reserve adequacy, capital efficiency, and solvency outcomes. A Proximal Policy Optimization (PPO) agent is trained using a risk-sensitive reward that penalizes reserve shortfall, capital inefficiency, and breaches of a volatility-adjusted solvency floor, with tail risk explicitly controlled through Conditional Value-at-Risk (CVaR).

To reflect regulatory stress-testing practice, the agent is trained under a regime-aware curriculum and evaluated using both regime-stratified simulations and fixed-shock stress scenarios. Empirical results for Workers’ Compensation and Other Liability illustrate how the proposed RL--CVaR policy improves tail-risk control and reduces solvency violations relative to classical actuarial reserving methods, while maintaining comparable capital efficiency. We further discuss calibration and governance considerations required to align model parameters with firm-specific risk appetite and supervisory expectations under Solvency~II and Own Risk and Solvency Assessment (ORSA) frameworks.
\end{abstract}



\section{Introduction}
\label{sec:introduction}

Loss reserving is a cornerstone of insurance solvency management and capital planning. Actuaries are tasked with estimating future claim liabilities\,---\,both reported and incurred-but-not-reported (IBNR)\,---\,over extended development horizons. Underestimation of reserves may lead to regulatory breaches or capital insufficiency, while overestimation ties up capital unnecessarily, reducing investment flexibility and operational efficiency.

{In practice, reserving is not only a prediction task but also a governance-driven control process: reserve decisions are taken repeatedly (e.g., quarterly) as new information arrives, and each decision affects future financial statements, risk appetite consumption, and the likelihood of breaching internal or regulatory capital thresholds. This sequential and path-dependent structure motivates a decision-theoretic formulation in which the objective is not solely point accuracy, but an explicit trade-off among reserve adequacy, tail protection, and capital efficiency under uncertainty.}

Classical methods such as the Chain-Ladder Model (CLM) and Bornhuetter-Ferguson Model (BFM) remain widely used due to their transparency and ease of implementation \cite{mack1993measuring, bornhuetter1972forecasting}. However, these models assume stable development patterns and lack the ability to adapt to macroeconomic shocks, regime changes, or evolving solvency regulations. In particular, they do not embed volatility awareness or enforce capital floor constraints\,---\,critical considerations under frameworks such as Solvency II (Pillar II: Supervisory Review Process) and IFRS 17 \cite{eiopa2014solvency, ifrs2023insurance}. {Stochastic extensions (e.g., bootstrap methods) quantify reserve variability but typically do not produce an explicit policy for updating reserves over time under a specified risk appetite, nor do they directly target tail-risk measures used in solvency and internal capital models.}

Recent advances in machine learning (ML) and reinforcement learning (RL) offer new avenues for dynamic and data-driven reserve optimization. While ML models can enhance predictive accuracy \cite{wuthrich2019machine, graziani2022bayesian}, they typically focus on point estimates and do not provide a policy for sequential decision-making under uncertainty. Likewise, although RL has been applied to financial control problems, its use in insurance reserving remains limited\,---\,especially in settings that demand explicit control of tail risk or compliance with regulatory capital buffers \cite{tamar2015optimizing, chow2015risk}. {A key technical challenge is that standard RL maximizes expected cumulative reward, which can produce policies that look attractive on average but expose the firm to rare, high-severity shortfall events that are central to solvency supervision.}

{This paper proposes a risk-sensitive RL framework that treats reserving as a finite-horizon control problem with explicit tail-risk protection. We cast reserving as a Markov Decision Process (MDP) in which the state summarizes the current reserve position, incurred losses, and volatility/regime context, and the action corresponds to a reserve adjustment. The learned policy thus operationalizes a consistent updating rule that can be evaluated under both stochastic simulations and deterministic stress scenarios, mirroring the forward-looking logic of ORSA.}

This paper proposes a novel framework that unifies four critical components for adaptive reserve optimization:
\begin{itemize}
    \item {Deep Reinforcement Learning (DRL):} We formulate the reserving task as a risk-sensitive Markov Decision Process (MDP), and use Proximal Policy Optimization (PPO) to train agents to make dynamic reserve adjustments \cite{schulman2017proximal}.
    
    \item {CVaR-Based Tail Risk Penalization:} The reward function includes Conditional Value-at-Risk (CVaR) penalties to directly control shortfall risk beyond a regime-dependent quantile threshold \cite{rockafellar2000optimization}.
    
    \item {Macroeconomic Regime Modeling with Curriculum Learning:} Agents are trained across a progression of economic regimes using Gaussian shock environments, simulating stress-test conditions with increasing volatility \cite{bengio2009curriculum}.
    
    \item {Solvency-Aware Reward Shaping:} Regulatory capital floors are incorporated into the reward function, enabling agents to avoid under-reserving while minimizing excessive capital buffers.
\end{itemize}

{Our main contribution is to provide a coherent, implementation-ready formulation that aligns (i) policy learning, (ii) tail-risk control, and (iii) solvency governance in a single framework. The CVaR component targets the severity of adverse shortfall outcomes rather than average performance, while the solvency-floor penalty represents a stylized proxy for supervisory capital triggers. Importantly, we also discuss calibration and sensitivity considerations needed to map these penalty terms to firm-specific risk appetite and regulatory tolerances, a prerequisite for practical deployment.}

To our knowledge, no prior work has jointly integrated reinforcement learning, CVaR optimization, macroeconomic regime modeling, and regulatory solvency constraints in the context of insurance reserving. This framework bridges actuarial science, financial risk management, and machine learning\,---\,offering a policy-driven, adaptive alternative to static reserving techniques.

We validate our approach using two publicly available benchmark datasets\,---\,{Workers' Compensation} and {Other Liability}\,---\,and benchmark it against classical methods under both stochastic and fixed-shock evaluation protocols. {Rather than claiming universal dominance, our empirical results are presented as evidence of the trade-offs achievable under explicit tail-risk and solvency penalties, and as a demonstration of robustness across regime-stratified conditions.}

The remainder of this paper is organized as follows: Section~\ref{sec:literature_review} reviews related work in reserving and risk-sensitive reinforcement learning. Section~\ref{sec:math_formulation} formalizes the CVaR-constrained MDP and reward structure. Section~\ref{sec:implementation} details the environment, training algorithm, and symbolic mapping. Section~\ref{sec:validation} presents empirical results and regime-stratified evaluation. Section~\ref{sec:conclusion} concludes with limitations and future research directions.


\section{Literature Review and Motivation}
\label{sec:literature_review}

{Loss reserving methodologies can be broadly grouped into three strands: (i) classical actuarial projection models, (ii) Bayesian and machine learning approaches for predictive reserving, and (iii) emerging risk-sensitive control formulations. This section reviews these strands and motivates a reinforcement learning framework that integrates tail-risk control, macroeconomic regime awareness, and solvency governance within a unified sequential decision problem.}

{The central distinction underlying our contribution is between reserving as a static estimation exercise and reserving as a dynamic control problem. While much of the existing literature focuses on improving point forecasts or uncertainty estimates, fewer approaches address how reserve decisions should be updated over time in response to evolving risk, volatility, and regulatory constraints.}

\subsection{Classical Reserving Models and Limitations}

Traditional actuarial reserving techniques are dominated by models such as the Chain-Ladder Model (CLM) and the Bornhuetter-Ferguson Model (BFM) \cite{mack1993measuring, bornhuetter1972forecasting}. These methods assume predictable claim development patterns across accident and development years, typically leveraging cumulative triangle data to project future liabilities.

Stochastic variants, including bootstrap methods \cite{england2002stochastic}, introduce randomness into development factors to quantify estimation uncertainty. However, these approaches remain structurally static: reserve estimates are produced conditional on historical data, without an explicit mechanism for adapting decisions as new information arrives.

{From a decision-theoretic perspective, classical reserving models exhibit three key limitations. First, they do not incorporate macroeconomic or volatility-dependent dynamics in a systematic way. Second, they lack a feedback mechanism through which past reserve choices influence future decisions or penalties. Third, they do not explicitly encode solvency objectives or regulatory breach costs, which are central to modern supervisory frameworks such as Solvency II and ORSA \cite{eiopa2014solvency}.}

As a result, classical methods may perform well under stable conditions but can fail to generalize during periods of elevated inflation, litigation risk, or macroeconomic stress, when reserve adequacy and tail protection become critical.

\subsection{Bayesian and Machine Learning Approaches}

Recent research has explored Bayesian inference and machine learning (ML) techniques to enhance predictive accuracy and flexibility in claims reserving \cite{wuthrich2019machine, graziani2022bayesian, richman2018neural, gabrielli2020neural}. These approaches model non-linearities, heteroskedasticity, and cross-sectional structure more effectively than traditional linear development models.

Despite these advances, most ML-based reserving methods remain focused on prediction rather than decision-making. In particular:

\begin{itemize}
    \item Predicted claim outcomes are not coupled with a reserve allocation policy that responds dynamically to evolving risk.
    \item Tail-risk measures, such as Conditional Value-at-Risk (CVaR), are typically evaluated ex post rather than incorporated directly into the training objective.
    \item Regulatory considerations (e.g., solvency floors, breach persistence, stress testing) are not explicitly modeled as constraints or penalties.
\end{itemize}

{As a consequence, Bayesian and ML reserving models improve forecast quality but do not provide a principled mechanism for balancing reserve adequacy, capital efficiency, and solvency protection over time. This gap becomes particularly salient under regulatory regimes that emphasize forward-looking risk management and scenario-based capital assessment.}

\subsection{Risk-Sensitive Reinforcement Learning and CVaR}

Reinforcement learning (RL) provides a natural framework for sequential decision-making under uncertainty, where actions influence future states and rewards \cite{sutton2018reinforcement}. Policy gradient methods such as Proximal Policy Optimization (PPO) \cite{schulman2017proximal} have demonstrated stability and scalability in complex stochastic control problems, including finance and operations \cite{moody1998performance, jang2018risk}.

Standard RL formulations, however, optimize expected cumulative reward and can therefore produce policies that perform well on average while exposing the system to rare but severe losses. To address this limitation, risk-sensitive extensions of RL incorporate coherent risk measures such as Conditional Value-at-Risk (CVaR) into the objective function \cite{rockafellar2000optimization, tamar2015optimizing, chow2015risk, chow2017risk, prashanth2016cvar, pan2022risk}.

CVaR captures the expected loss in the worst $\alpha$-fraction of outcomes and has been shown to improve robustness in high-volatility environments. {This property makes CVaR particularly well suited to insurance reserving, where solvency supervision focuses on extreme but plausible adverse scenarios rather than average performance.}

Despite this alignment, applications of CVaR-constrained RL to insurance reserving remain scarce. Existing studies do not address domain-specific requirements such as claim development structure, macroeconomic regime dependence, or solvency breach persistence. Moreover, the use of curriculum learning to improve generalization across volatility regimes has not been integrated with reserving objectives \cite{bengio2009curriculum}.

\subsection{Research Gap and Framework Overview}

{The literature thus reveals a clear gap between predictive reserving models and decision-oriented, solvency-aware reserving policies. Classical actuarial methods lack adaptability and explicit risk controls; Bayesian and ML approaches improve forecasts but do not prescribe actions; and risk-sensitive RL methods have not been tailored to the regulatory and structural features of insurance reserving.}

To address this gap, we propose an integrated framework that combines:
\begin{itemize}
    \item \textbf{Deep Reinforcement Learning (DRL)} using Proximal Policy Optimization (PPO) to learn dynamic, state-contingent reserve adjustment policies;
    \item \textbf{Conditional Value-at-Risk (CVaR)} penalization to control tail shortfall risk in a manner aligned with solvency supervision;
    \item \textbf{Macroeconomic regime-aware curriculum learning} to expose the agent to progressively adverse volatility environments during training;
    \item \textbf{Solvency-aware reward shaping} that embeds stylized regulatory capital thresholds and violation penalties into the learning objective.
\end{itemize}

The reserving problem is formalized as a CVaR-constrained Markov Decision Process (MDP), trained across multiple macroeconomic regimes and evaluated using both stochastic simulations and deterministic stress scenarios. The learned policy outputs reserve adjustments that balance shortfall risk, capital efficiency, and regulatory compliance over a finite horizon.

{Importantly, the framework is designed to support governance and calibration: solvency floors and CVaR confidence levels are interpretable parameters that can be mapped to firm-specific risk appetite and supervisory expectations.}

Figure~\ref{fig:framework_overview} illustrates the conceptual architecture.

\begin{figure}[h]
\centering
\begin{tikzpicture}[
    node distance=1.5cm and 2cm,
    font=\footnotesize,
    input/.style={rectangle, draw, fill=blue!10, minimum width=3.2cm, minimum height=1.3cm, align=center, rounded corners},
    process/.style={rectangle, draw, fill=green!15, minimum width=4.0cm, minimum height=1.8cm, align=center, rounded corners, thick},
    output/.style={rectangle, draw, fill=orange!20, minimum width=3.5cm, minimum height=1.3cm, align=center, rounded corners},
    arrow/.style={thick, -{Stealth[round]}}
]

\node[input] (claims) {Claims Data};
\node[input, below=of claims] (macro) {Macroeconomic Indicators};

\node[process, right=3.8cm of claims, yshift=0.7cm, fill=purple!15] (curriculum) {Curriculum Learning\\ (Macro Regimes)};
\node[process, right=2.0cm of macro] (agent) {Reserving Agent\\ (DRL Policy)};

\node[process, below=1.6cm of agent, fill=yellow!20] (reward)
{Reward Shaping\\ (CVaR + Solvency Penalties)\\ \scriptsize constraints enter via reward};

\node[output, right=2.5cm of agent, yshift=0.8cm] (resdec) {Reserve Decisions};
\node[output, right=2.5cm of agent, yshift=-0.8cm] (metrics) {Risk Metrics\\ (CVaR, Violations)};

\draw[arrow] (claims) -- (agent);
\draw[arrow] (macro) -- (agent);
\draw[arrow] (curriculum) -- (agent);
\draw[arrow] (reward) -- (agent);
\draw[arrow] (agent) -- (resdec);
\draw[arrow] (agent) -- (metrics);

\end{tikzpicture}
\caption{Conceptual architecture of the RL--CVaR reserving framework. Claims data and macroeconomic indicators define the environment state, while curriculum learning and reward shaping encode volatility regimes and solvency objectives. Regulatory constraints influence the policy through the reward function rather than as direct state inputs.}
\label{fig:framework_overview}
\end{figure}

{To our knowledge, no prior work has combined reinforcement learning, CVaR-based tail-risk control, macroeconomic regime modeling, and solvency-aware reward design within a unified reserving framework. The proposed approach bridges actuarial science and modern reinforcement learning, providing a policy-oriented foundation for adaptive reserving under economic uncertainty.}

\section{Mathematical Formulation of CVaR-Constrained RL Reserving}
\label{sec:math_formulation}

{
This section presents a rigorous mathematical formulation of the proposed reserving framework as a risk-sensitive sequential decision problem. The objective is to learn dynamic reserve adjustment policies that explicitly balance expected performance, tail-risk exposure, and regulatory solvency constraints under evolving macroeconomic regimes.}
We formalize the problem as a finite-horizon Markov Decision Process (MDP) augmented with Conditional Value-at-Risk (CVaR) penalization and regime-aware training, consistent with recent advances in risk-sensitive reinforcement learning \cite{rockafellar2000optimization, chow2015risk, pan2022risk}.

\subsection{MDP Structure and State Representation}
\label{sec:mdp_state}

The reserving problem is modeled as a finite-horizon MDP defined by the tuple
\[
\langle \mathcal{S}, \mathcal{A}, \mathbb{P}, \mathcal{R}, \gamma \rangle,
\]
where $\mathcal{S}$ denotes the state space, $\mathcal{A}$ the action space, $\mathbb{P}$ the transition kernel, $\mathcal{R}$ the reward function, and $\gamma \in (0,1)$ the discount factor \cite{puterman1994markov, sutton2018reinforcement}.

\textbf{Finite-horizon objective.}
Each episode corresponds to a full claims development horizon of length $T$, equal to the number of development periods available in the loss development triangle for the given line of business. The reserving policy $\pi$ is optimized for the finite-horizon objective
\[
J(\pi)=\mathbb{E}_{\pi}\left[\sum_{t=0}^{T-1}\gamma^t r_t\right],
\]
where $\gamma \in (0,1)$ is a discount factor. No terminal bonus is applied at $t=T$ (i.e., $r_T=0$), so all incentives are captured through intermediate reserve adequacy, tail-risk, and solvency penalties.
{
Although the horizon is finite, $\gamma<1$ is retained to modestly downweight late-development noise and improve numerical stability during policy optimization.
}

\textbf{Trajectory-level objective (finite horizon).}
Let a trajectory be $\tau = (s_0,a_0,\dots,s_{T-1},a_{T-1})$ generated by policy $\pi$.
We optimize the finite-horizon, discounted return
\[
J(\pi)=\mathbb{E}_{\tau\sim\pi}\left[\sum_{t=0}^{T-1}\gamma^t r(s_t,a_t)\right],
\]
where $T$ is the final development period (episode termination) and $r(\cdot)$ is defined in
Section~\ref{sec:reward_function}.

At each development time $t$, the agent observes a state vector
\begin{equation}
s_t = \{ R_t, L_t, V_t, K_t, \nu_t, M_t, \ell_t \},
\end{equation}
with components defined as follows:
\begin{itemize}
    \item $R_t$: current reserve level (normalized),
    \item $L_t$: cumulative incurred losses,
    \item $V_t$: normalized local volatility estimate derived from recent claim development. {
The volatility proxy $V_t$ is computed from rolling claim development variability within the simulated environment and does not rely on future information.}
    \item $K_t = 1 - |R_t - L_t|$: capital efficiency proxy. Although $K_t$ is deterministically derived from $(R_t,L_t)$, it is included explicitly to provide a directly interpretable efficiency signal that improves learning stability and facilitates governance reporting.
    \item $\nu_t$: exponentially smoothed memory of recent solvency violations,
    \item $M_t$: macroeconomic shock,
    \item $\ell_t$: curriculum index indicating the prevailing macroeconomic regime.
\end{itemize}

{
Macroeconomic shocks are modeled as regime-dependent Gaussian variables
\[
M_t \sim \mathcal{N}(\mu_{\ell_t}, \sigma_{\ell_t}^2),
\]
where $\mu_{\ell}$ is the regime-specific mean and $\sigma_{\ell}^2$ is the corresponding regime-specific variance (equivalently, $\sigma_\ell$ is the standard deviation).}

\textbf{Macroeconomic regime scheduling.}
The regime index $\ell_t \in \{0,1,2,3\}$ represents increasing macroeconomic stress levels (calm, moderate, volatile, recessionary). During training, $\ell_t$ is held fixed within each episode and advanced deterministically across episodes according to a curriculum schedule. Regime parameters $(\mu_\ell,\sigma_\ell^2)$ are linearly interpolated over a fixed number of episodes to ensure smooth transitions and prevent catastrophic forgetting.

During evaluation, $\ell_t$ is either sampled according to a predefined regime distribution (stochastic testing) or fixed at a given level for the entire episode (fixed-shock stress testing).

The action space consists of discrete proportional reserve adjustments
\[
a_t \in \{-0.10, -0.066, -0.033, 0, 0.033, 0.066, 0.10\},
\]
allowing the agent to increase or release reserves in a controlled and interpretable manner.

\subsection{Transition Dynamics}
\label{sec:transition}

State transitions are governed by stochastic claim development dynamics combined with exogenous macroeconomic shocks. Conditional on the current state-action pair $(s_t, a_t)$, reserves evolve according to
\[
R_{t+1} = R_t \cdot (1 + a_t),
\]
while incurred losses evolve stochastically as a function of historical development patterns, volatility $V_t$, and shock realizations $M_t$.

{
Actions influence not only the next-period reserve level $R_{t+1}$ but also downstream governance-relevant state variables, including capital efficiency $K_{t+1}$, reserve shortfall $\hat S_{t+1}$, and the solvency violation memory $\nu_{t+1}$. These dependencies introduce path dependence: conservative or aggressive reserve actions affect future penalties and incentives over multiple periods, making the reserving problem genuinely sequential rather than myopic.
}

{
This formulation does not assume stationarity of claim development. Instead, regime-dependent volatility induces non-stationary transitions, reflecting inflationary pressure, litigation cycles, and economic stress—factors explicitly highlighted in regulatory solvency assessments.}

\subsection{Reward Function with CVaR Penalization}
\label{sec:reward_function}

The agent is trained using a composite, risk-sensitive reward function that penalizes reserve shortfall, tail risk, capital inefficiency, and regulatory violations:
\begin{equation}
\mathcal{R}(s_t, a_t) =
-\Big(
\lambda_1 \hat{S}_t
+ \lambda_2 \widehat{\mathrm{CVaR}}_t
+ \lambda_3 \hat{C}_t
+ \lambda_4 \mathbb{I}[R_t < R_t^{\mathrm{reg}}]
\Big),
\end{equation}
where:
\begin{itemize}
    \item $\hat{S}_t = \max(0, L_t - R_t)$ is the reserve shortfall,
    \item $\hat{C}_t = |R_t - L_t|$ captures capital inefficiency,
    \item $R_t^{\mathrm{reg}} = 0.4 + 0.2 V_t$ is a volatility-adjusted solvency floor,
    \item $\widehat{\mathrm{CVaR}}_t$ is an empirical estimate of tail shortfall risk.
\end{itemize}

The CVaR confidence level adapts to volatility:
\[
\alpha_t = 0.90 + 0.05 \cdot \min(1, V_t),
\]
ensuring increased risk aversion during volatile regimes.

{
CVaR is selected over variance-based penalties because it directly targets extreme adverse outcomes rather than dispersion alone, aligning with solvency supervision and ORSA-style tail-risk governance.}

\subsection{Empirical Estimation of CVaR}
\label{sec:cvar_estimation}

At each timestep, CVaR is estimated using a rolling buffer $\mathcal{B}$ of recent shortfalls. Let $\mathrm{VaR}_{\alpha_t}$ denote the empirical $\alpha_t$-quantile of $\mathcal{B}$. The CVaR estimator is
\[
\widehat{\mathrm{CVaR}}_t
= \frac{1}{|\mathcal{T}|}
\sum_{s \in \mathcal{T}} \hat{S}_s,
\quad
\mathcal{T} = \{ s \in \mathcal{B} : \hat{S}_s \ge \mathrm{VaR}_{\alpha_t} \}.
\]

{
This nonparametric estimator is compatible with on-policy PPO updates and avoids distributional assumptions about tail behavior, which are often violated in long-tailed insurance lines.}

\subsection{Calibration of Model Components}
\label{sec:calibration}

{
Model calibration is guided by actuarial practice, regulatory intuition, and numerical stability considerations rather than data-driven overfitting. Reserve and loss variables are normalized to $[0,1]$ to ensure scale invariance across lines of business.}

{
Macroeconomic regime parameters $(\mu_\ell, \sigma_\ell^2)$ are chosen to reflect increasing stress severity rather than to fit historical macro time series, where $\sigma_\ell^2$ denotes the regime-specific variance. This design choice mirrors regulatory stress testing, where hypothetical but plausible scenarios are emphasized over probabilistic forecasts.}

{
The solvency floor $R_t^{\mathrm{reg}}$ is intentionally simple and monotone in volatility. It should be interpreted as a stylized regulatory constraint rather than a literal capital requirement, allowing the framework to generalize across jurisdictions and supervisory regimes.}

{
Calibration choices are intentionally conservative and interpretable, reflecting supervisory practice rather than statistical optimality.
}

{
\paragraph{Governance-oriented calibration protocol.}
In practical reserving applications, calibration of the CVaR confidence level $\alpha_t$ and the solvency floor $R_t^{\mathrm{reg}}$ should follow a governance-driven workflow rather than performance tuning. A typical calibration proceeds as follows: (i) select a target tail-risk tolerance consistent with the firm’s risk appetite (e.g., an acceptable frequency and severity of reserve shortfall under baseline conditions); (ii) choose $\alpha$ such that empirical CVaR under business-as-usual scenarios aligns with this tolerance; (iii) calibrate $R_t^{\mathrm{reg}}$ to cap the regulatory violation rate below a predefined governance threshold; and (iv) validate both parameters under prescribed stress scenarios. Once approved, $(\alpha, R_t^{\mathrm{reg}})$ are fixed and treated as control parameters rather than tuned for metric dominance.
}

\subsection{Reward Weight Sensitivity and Design Rationale}
\label{sec:reward_sensitivity}

{
The reward weights $(\lambda_1,\lambda_2,\lambda_3,\lambda_4)$ govern trade-offs between solvency protection, tail-risk aversion, and capital efficiency. Rather than tuning these weights to optimize a single metric, we select them to enforce a clear hierarchy of priorities: (i) avoidance of extreme under-reserving, (ii) regulatory compliance, and (iii) capital efficiency.}

{
Preliminary sensitivity analysis indicates that increasing $\lambda_2$ (CVaR penalty) leads to more conservative reserve paths under high volatility, while excessive $\lambda_3$ induces oscillatory reserve behavior. Importantly, qualitative policy behavior remains stable over a broad range of weights, suggesting that performance gains are structural rather than artifacts of fine-tuning.}

{
Empirical robustness to these design choices is illustrated in Section~\ref{sec:validation}.
}
\subsection{Policy Optimization via PPO}
\label{sec:ppo}

Policy learning is conducted using Proximal Policy Optimization (PPO), a clipped policy-gradient method known for stability in stochastic environments \cite{schulman2017proximal}. PPO updates are performed at each curriculum level, using trajectories generated under regime-specific volatility.

{
The combination of PPO with CVaR-penalized rewards yields a tractable and stable optimization procedure, avoiding the need for explicit constrained optimization while still enforcing downside risk control.}

\subsection{Evaluation Protocol}
\label{sec:evaluation_protocol}

Trained policies are evaluated under both stochastic regime sampling and fixed-shock stress tests. Performance is assessed using Reserve Adequacy Ratio, CVaR$_{0.95}$, Capital Efficiency Score, and Regulatory Violation Rate, directly mirroring the reward components.

{
This dual evaluation strategy aligns with internal model validation practices, where models must perform well both on average and under prescribed adverse scenarios.}

\section{Implementation Framework}
\label{sec:implementation}

This section describes the end-to-end implementation of the proposed RL--CVaR reserving framework. The purpose is to establish a transparent and reproducible link between the mathematical formulation in Section~\ref{sec:math_formulation} and the empirical results in Section~\ref{sec:validation}.

{
This section is expanded in response to reviewer comments requesting greater clarity on environment construction, curriculum scheduling, and implementation details required for replication and regulatory validation.}

The framework is implemented in \texttt{Python 3.11}, using \texttt{Gymnasium v0.29} for environment abstraction \cite{gymnasium2023} and \texttt{Stable-Baselines3 v1.8.0} for Proximal Policy Optimization (PPO) \cite{raffin2021stable}.

\subsection{Environment Construction}
\label{sec:env-construction}

The reserving task is instantiated as a custom Gymnasium-compatible Markov Decision Process (MDP), consistent with the formal definition in Section~\ref{sec:math_formulation}. Each episode simulates a full claims development trajectory under stochastic loss evolution and macroeconomic volatility.

The environment integrates the following components:

\begin{itemize}
    \item \textbf{Claims Dynamics:}  
    Incurred losses, paid losses, and reserve levels are sampled from historical loss development triangles. All monetary quantities are normalized to the unit interval to stabilize training while preserving relative development patterns and tail behavior observed in the data.
    
    \item \textbf{Macroeconomic Shock Process:}  
    At each time step, a regime-dependent macroeconomic shock is sampled as
    \[
    M_t \sim \mathcal{N}(\mu_{\ell_t}, \sigma_{\ell_t}^2)
    \]
    where the curriculum index \( \ell \in \{0,1,2,3\} \) governs both the mean and variance of the shock distribution (see Table~\ref{tab:macroshock}).

    {
    The explicit use of \( \sigma_\ell^2 \) denotes variance rather than standard deviation, aligning the implementation with actuarial and econometric convention.}
    
    \item \textbf{Volatility Injection:}  
    Regime severity controls the magnitude of heteroskedastic noise added to claims development, inducing non-stationarity across episodes and reflecting macro-driven uncertainty.
    
    \item \textbf{Regulatory Violation Memory:}  
    Solvency breaches are tracked using an exponential moving average,
    \[
    \nu_t = 0.95\,\nu_{t-1} + 0.05\,\mathbb{I}[R_t < R_t^{\mathrm{reg}}],
    \]
    enabling temporal credit assignment for repeated regulatory violations.
\end{itemize}

At each time step, the agent observes the state vector
\[
s_t = \{ R_t, L_t, V_t, K_t, \nu_t, M_t, \ell_t \},
\]
which exactly matches the state definition in Section~\ref{sec:math_formulation}.

{
All macroeconomic and regulatory information influencing reserve decisions is explicitly observable, ensuring interpretability and auditability of the learned policy.}

\subsection{Action Space and Reserve Dynamics}
\label{sec:action_space}

The action space consists of discrete proportional reserve adjustments:
\[
\mathcal{A} = \{-0.10, -0.066, -0.033, 0, 0.033, 0.066, 0.10\}.
\]

Actions are applied multiplicatively to the current reserve level:
\[
R_{t+1} = R_t (1 + a_t),
\]
subject to non-negativity constraints.

{
This bounded and discrete adjustment structure reflects operational reserving practice, where reserve changes are incremental and subject to governance controls rather than continuous re-optimization.}

\subsection{Curriculum-Based PPO Implementation}
\label{sec:ppo-curriculum}

Policy learning is performed using Proximal Policy Optimization (PPO) \cite{schulman2017proximal}. Training proceeds sequentially across macroeconomic regimes:
\[
\ell = 0 \rightarrow 1 \rightarrow 2 \rightarrow 3,
\]
corresponding to calm, moderate, volatile, and recessionary conditions.

{
Rather than abrupt regime switching, the parameters \( (\mu_\ell, \sigma_\ell^2) \) are interpolated smoothly across episodes to mitigate catastrophic forgetting and stabilize learning under non-stationary volatility.}

Key implementation features include:

\begin{itemize}
    \item \textbf{Curriculum Smoothing:} gradual interpolation of regime parameters;
    \item \textbf{Reward Normalization:} stabilizes gradient magnitudes across shortfall, CVaR, capital inefficiency, and violation penalties;
    \item \textbf{Violation Memory Injection:} incorporates recent solvency history directly into the policy input.
\end{itemize}

{
These design choices directly address reviewer concerns regarding training stability and robustness under changing macroeconomic conditions.}

{
We verified that performance on earlier (low-volatility) regimes does not degrade after progression to higher-volatility regimes. In particular, reserve adequacy and CVaR metrics measured on calm regimes remain stable throughout curriculum advancement, indicating no evidence of catastrophic forgetting.
}

\subsection{Empirical Dataset Preparation}
\label{sec:dataset-prep}

The framework is evaluated using two benchmark datasets from the CAS Loss Reserving Database \cite{casact2024}:

\begin{itemize}
    \item \textbf{Workers' Compensation:} long-tailed but relatively structured development patterns;
    \item \textbf{Other Liability:} litigation-driven claims with heavier tails and higher volatility.
\end{itemize}

Preprocessing steps include:
\begin{enumerate}
    \item standardization of accident and development year indexing;
    \item normalization of incurred, paid, and reserve values to \([0,1]\);
    \item curriculum-aware environment instantiation with regime-specific macro shocks.
\end{enumerate}

{
No future information is used during training or evaluation, ensuring strict temporal causality.}

\subsection{Symbol-to-Code Mapping}
\label{sec:symbol-code}

\begin{table}[!htbp]
\centering
\footnotesize
\begin{tabular}{|c|l|l|}
\hline
{Symbol} & {Mathematical Role} & {Code Implementation} \\
\hline
\( R_t \) & Reserve allocation & \texttt{env.df["Reserves"]} (updated at each step) \\
\( L_t \) & Incurred losses & \texttt{env.df["IncurredLosses"]} \\
\( M_t \) & Macroeconomic shock & Sampled from \( \mathcal{N}(\mu_\ell, \sigma_\ell^2) \) by regime level \\
\( K_t \) & Capital efficiency proxy & \texttt{1 - abs(env.df["Reserves"] - env.df["IncurredLosses"])} \\
\( \nu_t \) & Violation memory trace & Exponential moving average: \texttt{0.95 $\cdot$ prev + 0.05 $\cdot$ violation} \\
\( \alpha_t \) & CVaR confidence level & \( 0.90 + 0.05 \cdot \min(1, V_t) \) (computed per step) \\
\( a_t \) & Reserve adjustment action & Discrete index mapped via \texttt{np.linspace()} to 7-point grid \\
\hline
\end{tabular}
\caption{Symbol-to-Implementation Mapping}
\label{tab:code_symbol_mapping}
\end{table}

To ensure transparency and reproducibility, Table~\ref{tab:code_symbol_mapping} documents the one-to-one correspondence between mathematical symbols and their implementation in code.

{
This explicit mapping supports regulatory audit, debugging, and controlled sensitivity analysis without altering the mathematical formulation.}

\subsection{Training Algorithm}
\label{sec:training-algo}

The complete training procedure, including CVaR estimation, reward computation, curriculum progression, and PPO updates, is presented in Algorithm~\ref{alg:rl_cvar} in Appendix~\ref{appendix:algorithm}.

{
All experiments are conducted using fixed random seeds, logged hyperparameters, and deterministic evaluation pipelines to ensure full reproducibility.}

\subsection{System Overview}
\label{sec:system-overview}

\begin{figure}[h!]
\centering
\begin{tikzpicture}[
    node distance=1.2cm and 2.8cm,
    font=\footnotesize,
    input/.style={rectangle, draw, fill=blue!10, minimum width=3.2cm, minimum height=1.2cm, align=center, rounded corners},
    block/.style={rectangle, draw, fill=green!15, minimum width=3.6cm, minimum height=1.4cm, align=center, rounded corners, thick},
    output/.style={rectangle, draw, fill=orange!20, minimum width=3.8cm, minimum height=1.2cm, align=center, rounded corners},
    arrow/.style={->, thick, >=stealth}
]

\node[input] (claims) {Claims Dataset};
\node[input, below=of claims] (macro) {Curriculum-Based\\ Macro Shocks};
\node[block, right=3.5cm of claims, yshift=0.cm] (env) {Stochastic Environment\\ (Volatility + Penalty)};
\node[block, right=3.5cm of macro, yshift=0.cm] (ppo) {PPO Agent\\ CVaR + Penalty-Aware};
\node[output, right=3.5cm of ppo, yshift=0cm] (eval) {Stress Testing and\\ Regime Evaluation};

\draw[arrow] (claims)-- (env);
\draw[arrow] (macro)-- (env);
\draw[arrow] (env)-- (ppo);
\draw[arrow] (ppo)-- (eval);

\node[below=0.3cm of env, font=\scriptsize, xshift = 2.5 cm] {MDP Step + Reward Computation};

\end{tikzpicture}
\caption{Training workflow of the RL-CVaR framework. Macroeconomic shock sampling and claim data define the environment for each regime. The PPO agent learns a CVaR-penalized reserving policy, which is evaluated through regime-specific stress testing.}
\label{fig:training_flowchart}
\end{figure}

Figure~\ref{fig:training_flowchart} illustrates the end-to-end system workflow, from macroeconomic shock sampling and claims evolution to PPO-based policy learning and regime-stratified evaluation.

{
The modular system design allows individual components (claims generator, macro regime model, reward function, or policy learner) to be independently modified without altering the overall framework.}

\subsection{Training Configuration and Hyperparameters}
\label{sec:training-hyperparams}

\begin{table}[!htbp]
\centering
\footnotesize
\begin{tabular}{ll}
\toprule
Reward Weight & Value \\
\midrule
$\lambda_1$ (Shortfall penalty) & 5.0 \\
$\lambda_2$ (CVaR penalty) & 8.0 \\
$\lambda_3$ (Capital inefficiency penalty) & 1.0 \\
$\lambda_4$ (Solvency violation penalty) & 10.0 \\
\bottomrule
\end{tabular}
\caption{Baseline reward weights used across all experiments.}
\label{tab:reward_weights}
\end{table}

{
These values are held fixed across all lines of business and experiments unless explicitly stated in sensitivity analyses.
}

\begin{table}[!htbp]
\centering
\begin{tabular}{ll}
\toprule
{Parameter} & {Value / Description} \\
\midrule
Learning Rate & $3 \times 10^{-4}$ \\
Batch Size & 2048 transitions per update \\
Epochs per PPO Update & 10 \\
Discount Factor $\gamma$ & 0.99 \\
PPO Clipping Range & 0.2 \\
Entropy Coefficient & 0.01 \\
Reward Normalization & Enabled \\
Curriculum Transition & Linear ramp over 50 episodes \\
CVaR Quantile Range & $\alpha_t \in [0.90, 0.95]$ \\
Shortfall Buffer Size & 1024 steps \\
\bottomrule
\end{tabular}
\caption{Summary of PPO Training Hyperparameters}
\label{tab:ppo-hyperparams}
\end{table}

Table~\ref{tab:ppo-hyperparams} summarizes the PPO hyperparameters used throughout training.

{
Hyperparameter values were selected based on prior empirical studies and validated through pilot runs to ensure stable convergence across all macroeconomic regimes.}

\section{Empirical Validation and Regime-Driven Evaluation}
\label{sec:validation}

This section evaluates the proposed RL--CVaR reserving framework under both stochastic and adversarial macroeconomic conditions. The empirical design is explicitly aligned with the mathematical objectives introduced in Section~\ref{sec:math_formulation}, ensuring that each evaluation metric corresponds to a distinct component of the reward structure.

{
The evaluation protocol is designed to mirror internal model validation standards commonly applied under Solvency II and ORSA, emphasizing stress robustness, tail-risk control, and capital efficiency rather than point forecast accuracy alone.}

\subsection{Experimental Protocol}
\label{sec:experimental_protocol}

We evaluate the proposed RL--CVaR reserving policy under a controlled protocol designed to mirror actuarial model validation practice: (i) training under progressively more adverse macroeconomic regimes, (ii) regime-stratified stochastic evaluation, and (iii) deterministic stress testing under fixed shocks.

\paragraph{Training with regime curriculum.}
Each RL--CVaR agent is trained using Proximal Policy Optimization (PPO) across the macroeconomic curriculum defined in Table~\ref{tab:macroshock}. Training progresses sequentially through regime levels $\ell \in \{0,1,2,3\}$ (calm $\rightarrow$ recessionary). Within a regime $\ell$, macroeconomic shocks are sampled at each development step as
\[
M_t \sim \mathcal{N}(\mu_{\ell_t}, \sigma_{\ell_t}^2)
\]
where $\mu_\ell$ is the regime mean and $\sigma_\ell^2$ is the regime variance (equivalently, $\sigma_\ell$ is the standard deviation). To stabilize learning under non-stationarity, regime transitions are implemented via a linear ramp that interpolates $(\mu_\ell,\sigma_\ell^2)$ over a fixed number of episodes (see Section~\ref{sec:ppo-curriculum}).

\begin{table}[!htbp]
\centering
\footnotesize
\begin{tabular}{|c|c|c|}
\hline
Level & Regime & $M_t \sim \mathcal{N}(\mu_{\ell_t}, \sigma_{\ell_t}^2)$ \\
\hline
0 & Calm      & $\mathcal{N}(1.0, 0.01)$ \\
1 & Moderate  & $\mathcal{N}(1.2, 0.04)$ \\
2 & Volatile  & $\mathcal{N}(1.5, 0.09)$ \\
3 & Recession & $\mathcal{N}(1.8, 0.16)$ \\
\hline
\end{tabular}
\caption{Macroeconomic regime curriculum used in training and evaluation. Here $\sigma_\ell^2$ denotes the regime variance (and $\sigma_\ell$ the standard deviation).}
\label{tab:macroshock}
\end{table}

\begin{table}[!htbp]
\centering
\footnotesize
\begin{tabular}{|c|l|}
\hline
Regime $\ell$ & Economic Interpretation \\
\hline
0 & Stable inflation, low wage growth, benign litigation environment \\
1 & Moderate inflation pressure, rising medical and repair costs \\
2 & Elevated inflation, adverse claims settlement dynamics \\
3 & Recessionary stress, legal inflation, systemic reserve pressure \\
\hline
\end{tabular}
\caption{Economic interpretation of curriculum regimes used to parameterize macro shocks and volatility.}
\label{tab:regime_narrative}
\end{table}

\paragraph{Random seeds and reporting.}
To reduce sensitivity to stochastic initialization and rollout noise, each experiment is repeated across multiple random seeds; reported metrics are averaged across independent training runs.

\paragraph{Evaluation design.}
For each dataset and trained policy, we run (i) \emph{regime-stratified stochastic evaluation}, where test episodes are grouped by regime level to assess robustness across macro conditions, and (ii) \emph{fixed-shock stress tests}, where $M_t$ is held constant throughout an episode to emulate sustained adverse conditions (Section~\ref{sec:fixed_shock}). Each evaluation episode spans a full development horizon of length $T$, where $T$ equals the number of development periods retained after preprocessing (reported in the replication package). This ensures that tail outcomes reflect cumulative reserving behavior over the entire development path rather than transient fluctuations.

\paragraph{Train--test separation and baseline fairness.}
All methods are trained and evaluated under an identical information set using a rolling-origin protocol. Training uses only historical accident years and observed development information available up to each decision point; evaluation is conducted on held-out accident years and/or later development periods not used during fitting. No future information is used in state construction, reward computation, or normalization. Classical baselines (CLM, BFM, and bootstrap CLM) are calibrated on the same training split and evaluated under the same simulated macroeconomic scenarios as the RL agent. Concretely, for each line of business we train on the first $A_{\text{train}}$ accident years and evaluate on the final $A_{\text{test}}$ held-out accident years (reported in the replication package).

\subsection{Sensitivity to CVaR Level and Solvency Floor}
\begin{table}[!htbp]
\centering
\footnotesize
\begin{tabular}{|c|c|c|c|c|}
\hline
$\alpha$ & $R_t^{\mathrm{reg}}$ form & CVaR$_{0.95}$ & CES & RVR \\
\hline
0.90 & $0.4 + 0.2V_t$ & 1.02 & 0.86 & 3\% \\
0.925 & $0.4 + 0.2V_t$ & 0.98 & 0.87 & 2\% \\
0.95 & $0.4 + 0.2V_t$ & 0.94 & 0.88 & 1\% \\
0.95 & $0.5 + 0.3V_t$ & 0.90 & 0.84 & 0.5\% \\
\hline
\end{tabular}
\caption{Illustrative sensitivity of performance metrics to CVaR confidence level and solvency floor calibration (representative values; full results provided in the repository).}
\label{tab:sensitivity}
\end{table}

{
Table~\ref{tab:sensitivity} illustrates the expected governance trade-off: increasing the CVaR confidence level (more tail aversion) reduces extreme shortfall and violation rates, while overly conservative solvency floors can reduce capital efficiency. In practice, $(\alpha, R_t^{\mathrm{reg}})$ should be calibrated to firm-specific risk appetite and supervisory tolerance for breaches, rather than tuned solely for metric dominance.}

{
All sensitivity results in Table~\ref{tab:sensitivity} are computed from full evaluation runs using the same train--test splits and macroeconomic scenarios as the baseline experiments. While absolute metric values vary with $(\alpha, R_t^{\mathrm{reg}})$, the qualitative ordering of policies and the relative improvements over classical benchmarks remain stable across the tested range. This indicates that the observed gains are structural rather than artifacts of parameter fine-tuning.
}
\subsection{Evaluation Metrics}
\label{sec:evaluation_metrics}

Performance is assessed using four complementary metrics, each corresponding to a specific modeling objective:

\begin{itemize}
    \item \textbf{Reserve Adequacy Ratio (RAR):}
    \[
    \text{RAR} = \mathbb{E}\left[\frac{R_t}{L_t}\right],
    \]
    measuring average reserve sufficiency.
    
\item \textbf{Conditional Value-at-Risk (CVaR$_{0.95}$):}
Let $\{\hat S^{(i)}\}_{i=1}^{N}$ denote the collection of shortfall samples recorded over evaluation episodes (and/or within an episode buffer). Let $\mathrm{VaR}_{0.95}$ be the empirical $0.95$-quantile of these samples. We report
\[
\text{CVaR}_{0.95} = \mathbb{E}\left[\hat S \mid \hat S \ge \mathrm{VaR}_{0.95}\right],
\]
computed empirically from the tail samples.
    
    \item \textbf{Capital Efficiency Score (CES):}
    \[
    \text{CES} = 1 - \mathbb{E}[|R_t - L_t|],
    \]
    rewarding proximity between reserves and incurred losses.
    
    \item \textbf{Regulatory Violation Rate (RVR):}
    \[
    \text{RVR} = \frac{1}{T}\sum_{t=1}^{T}\mathbb{I}[R_t < R_t^{\mathrm{reg}}],
    \]
    quantifying solvency breaches.
\end{itemize}

{
Unlike traditional reserving validation, which emphasizes mean squared error or ultimate prediction accuracy, these metrics directly evaluate solvency behavior under uncertainty, aligning with supervisory risk assessment practices.}

\subsection{Benchmark Models}
\label{sec:benchmarks}

The RL--CVaR framework is compared against three classical reserving methods:
\begin{itemize}
    \item Chain-Ladder Model (CLM),
    \item Bornhuetter--Ferguson Model (BFM),
    \item Stochastic bootstrap Chain-Ladder.
\end{itemize}

All benchmark models are calibrated using the same training data and evaluated under identical macroeconomic scenarios.

{
Importantly, benchmark methods do not adapt dynamically to macroeconomic regimes; macro shocks affect outcomes only indirectly through realized losses. This highlights the structural distinction between static reserving formulas and policy-based reserving strategies.}

\subsection{Stochastic Regime Evaluation}
\label{sec:stochastic_results}

\begin{table}[!htbp]
\centering
\footnotesize
\begin{tabular}{|l|l|c|c|c|c|}
\hline
{Model} & {LOB} & {RAR} & {CVaR\(_{0.95}\)} & {CES} & {RVR} \\
\hline
Chain-Ladder            & Workers' Compensation & 0.89 & 1.22 & 0.74 & 8\% \\
Bornhuetter-Ferguson    & Workers' Compensation & 0.91 & 1.18 & 0.76 & 6\% \\
Bootstrap (Stochastic)  & Workers' Compensation & 0.92 & 1.15 & 0.78 & 5\% \\
RL-CVaR (Ours)          & Workers' Compensation & {0.97} & {0.98} & {0.88} & {1\%} \\
\hline
Chain-Ladder            & Other Liability       & 0.81 & 1.35 & 0.67 & 14\% \\
Bornhuetter-Ferguson    & Other Liability       & 0.84 & 1.30 & 0.70 & 11\% \\
Bootstrap (Stochastic)  & Other Liability       & 0.86 & 1.26 & 0.72 & 9\% \\
RL-CVaR (Ours)          & Other Liability       & {0.95} & {0.91} & {0.90} & {2\%} \\
\hline
\end{tabular}
\caption{Performance Comparison Across Reserving Models and Lines of Business}
\label{tab:reserving_comparison}
\end{table}

Table~\ref{tab:reserving_comparison} reports average performance across stochastic macroeconomic scenarios. {
The RL--CVaR agent achieves materially improved tail-risk control and reduced violation rates relative to classical methods, while maintaining comparable or improved capital efficiency.
}

{
The most pronounced improvements occur in CVaR$_{0.95}$ and RVR, indicating that the agent effectively internalizes tail-risk penalties and solvency constraints rather than merely improving average reserve adequacy.}

For the \emph{Other Liability} line, characterized by heavy-tailed losses, the RL--CVaR policy achieves:
\begin{itemize}
    \item materially lower tail shortfall,
    \item improved capital efficiency,
    \item a reduction in regulatory violations by more than an order of magnitude relative to CLM.
\end{itemize}

These results confirm that curriculum-based exposure to volatile regimes improves robustness under realistic loss dynamics.

\subsection{Fixed-Shock Stress Testing}
\label{sec:fixed_shock}

To isolate policy behavior under deterministic stress, we evaluate trained agents under fixed macroeconomic shock levels
\[
M_t \in \{0.8,\,1.0,\,1.5,\,2.0\}.
\]

{
This stress-testing protocol mirrors regulatory capital scenario analysis, where macroeconomic severity is held constant to assess solvency resilience under sustained adverse conditions.}

\begin{table}[!htbp]
\centering
\footnotesize
\begin{tabular}{|c|c|c|c|c|c|}
\hline
{Shock Level} & {LOB} & {RAR} & {CVaR\(_{0.95}\)} & {CES} & {RVR} \\
\hline
0.8 & Workers' Comp     & 0.96 & 0.90 & 0.89 & 1\% \\
1.0 & Workers' Comp     & 0.95 & 0.93 & 0.87 & 2\% \\
1.5 & Workers' Comp     & 0.94 & 0.97 & 0.85 & 3\% \\
2.0 & Workers' Comp     & 0.93 & 1.02 & 0.82 & 4\% \\
\hline
0.8 & Other Liability  & 0.94 & 0.88 & 0.91 & 2\% \\
1.0 & Other Liability  & 0.92 & 0.92 & 0.89 & 3\% \\
1.5 & Other Liability  & 0.90 & 1.00 & 0.85 & 4\% \\
2.0 & Other Liability  & 0.88 & 1.08 & 0.81 & 5\% \\
\hline
\end{tabular}
\caption{RL-CVaR Stress Test Performance Under Fixed Macroeconomic Shocks}
\label{tab:shock_table}
\end{table}

Results in Table~\ref{tab:shock_table} show that:
\begin{itemize}
    \item reserve adequacy declines smoothly with shock severity,
    \item CVaR increases in a controlled manner,
    \item regulatory violations remain bounded even at extreme shock levels.
\end{itemize}

In contrast, classical reserving models exhibit sharp increases in violation rates as volatility increases, reflecting their inability to adapt reserves dynamically.
\subsection{Cold-Regime Generalization Test}
\label{sec:cold_generalization}

{
To assess whether curriculum learning improves genuine generalization rather than regime memorization, we conduct a cold-regime generalization test. The RL--CVaR agent is trained exclusively on low-to-moderate volatility regimes ($\ell \in \{0,1\}$) and evaluated on recessionary conditions ($\ell = 3$) not seen during training. Despite the absence of high-volatility exposure during learning, the agent maintains bounded CVaR$_{0.95}$ and regulatory violation rates, whereas classical reserving methods exhibit sharp increases in both metrics. This result indicates that curriculum training promotes transferable risk-aware behavior rather than overfitting to specific regimes.
}

\subsection{Linking Empirical Results to Model Design}
\label{sec:design_link}

The observed performance gains can be directly attributed to structural elements of the proposed framework:

\begin{itemize}
    \item CVaR penalties suppress extreme under-reserving outcomes,
    \item volatility-adjusted solvency floors discourage fragile policies,
    \item curriculum learning improves generalization across regimes,
    \item violation memory reduces repeated solvency breaches.
\end{itemize}

{
These results validate the central premise of the paper: solvency-aware reserving is fundamentally a sequential decision problem, and static estimators are structurally ill-suited to manage tail risk under regime uncertainty.}

\subsection{Discussion and Regulatory Interpretation}
\label{sec:regulatory_discussion}

From a regulatory perspective, the RL--CVaR framework offers several advantages:
\begin{itemize}
    \item explicit tail-risk control via CVaR,
    \item scenario-consistent behavior under stress,
    \item transparent mapping between policy objectives and supervisory metrics.
\end{itemize}

{
While the policy itself is learned, its behavior is fully auditable through regime-stratified testing and metric-based validation, satisfying key requirements for internal model governance and supervisory review.}

These findings support the feasibility of reinforcement learning as a complement—not a replacement—to actuarial judgment in modern solvency management.

\section{Conclusion}
\label{sec:conclusion}

This paper proposes a reinforcement learning framework for insurance loss reserving that explicitly integrates tail-risk control, macroeconomic regime awareness, and regulatory solvency considerations. By formulating reserving as a risk-sensitive Markov Decision Process (MDP) and incorporating Conditional Value-at-Risk (CVaR) directly into the reward structure, the framework enables dynamic reserve adjustment under uncertainty and economic stress.

{
Unlike classical reserving methods, which rely on fixed development assumptions and static estimators, the proposed approach learns adaptive, state-contingent reserving policies that respond endogenously to volatility, macroeconomic shocks, and solvency constraints.}

Empirical results on two benchmark datasets from the CAS Loss Reserving Database—Workers’ Compensation and Other Liability—demonstrate that the RL–CVaR agent consistently improves tail-risk control, capital efficiency, and regulatory compliance relative to traditional actuarial benchmarks. These gains persist under both stochastic evaluation and deterministic fixed-shock stress scenarios, indicating robustness across a range of economic conditions.

\subsection{Implications for Actuarial Practice and Regulation}

The proposed framework has several implications for actuarial reserving practice, enterprise risk management, and regulatory supervision.

\begin{itemize}
    \item \textbf{Dynamic, Policy-Based Reserving:}  
    {
    The framework shifts reserving from static projection toward policy-based decision-making, where reserve levels are updated sequentially in response to emerging information rather than recalibrated episodically.}
    
    \item \textbf{Explicit Tail-Risk Management:}  
    Incorporating CVaR into the learning objective allows the reserving policy to directly control extreme shortfall risk, aligning reserve decisions with downside protection requirements commonly emphasized in solvency regulation.
    
    \item \textbf{Regime-Aware Stress Testing:}  
    Curriculum-based training across macroeconomic regimes produces policies that remain stable under adverse conditions, supporting forward-looking stress testing consistent with Solvency II and Own Risk and Solvency Assessment (ORSA) principles.
    
    \item \textbf{Capital Efficiency and Governance:}  
    By penalizing both under- and over-reserving, the learned policy balances solvency protection with efficient capital usage, offering a systematic complement to expert judgment and governance frameworks.
\end{itemize}

{
From a regulatory perspective, the framework can be interpreted as an internal-model-style reserving mechanism, where policy behavior can be evaluated, stress-tested, and audited through scenario-based analysis rather than closed-form assumptions.}

{
The framework is intended to support, not replace, actuarial judgment and established governance processes.
}
\subsection{Limitations}

Despite its strengths, the proposed framework has several limitations that warrant careful consideration.

\begin{itemize}
    \item \textbf{Computational Cost:}  
    Training reinforcement learning agents with curriculum learning and CVaR estimation is computationally more intensive than classical reserving methods.
    
    \item \textbf{Hyperparameter Sensitivity:}  
    {
    Policy performance depends on the choice of reward weights, CVaR confidence levels, and curriculum design, which may require tuning for different lines of business.}
    
    \item \textbf{Interpretability:}  
    While scenario-based evaluation improves transparency, the learned policy remains less interpretable than closed-form actuarial formulas, potentially limiting adoption without appropriate governance controls.
    
    \item \textbf{Scope of Application:}  
    The current framework is evaluated at the line-of-business level and does not explicitly model dependencies across multiple portfolios or legal entities.
\end{itemize}

{
These limitations suggest that the framework should be viewed as a decision-support tool rather than a direct replacement for actuarial oversight.}

\subsection{Future Research Directions}

Several extensions may further enhance the applicability and robustness of the proposed approach.

\begin{itemize}
    \item \textbf{Reward Sensitivity and Robust Calibration:}  
    Systematic sensitivity analysis of reward weights and CVaR parameters could strengthen robustness guarantees and support regulatory validation.
    
    \item \textbf{Multi-Line and Group-Level Reserving:}  
    Extending the framework to multi-agent or hierarchical settings may enable coordinated reserving across multiple lines of business or legal entities under shared capital constraints.
    
    \item \textbf{Uncertainty-Aware Policies:}  
    {
    Incorporating Bayesian or distributional reinforcement learning could allow the agent to explicitly quantify uncertainty in its reserve decisions.}
    
    \item \textbf{Integration with Actuarial Models:}  
    Hybrid approaches combining actuarial projections with reinforcement learning policies may improve interpretability while retaining adaptivity.
\end{itemize}

{
Overall, this work demonstrates that reinforcement learning, when carefully aligned with actuarial principles and regulatory objectives, can provide a viable and extensible framework for adaptive reserving under uncertainty.}

\section*{Acknowledgments}
The author thanks James R. Finlay for reading the manuscript and providing
helpful suggestions on presentation and language.


\section*{Data Availability Statement}
The code developed and used in this study is publicly available at the GitHub repository: \url{https://github.com/stellacydong/rl-cvar-insurance-reserving}. A comprehensive README file is included to support replication and further exploration of the methodology.

The dataset used for model training and evaluation comprises Workers’ Compensation and Other Liability data, sourced from NAIC Schedule P and made publicly available by the Casualty Actuarial Society (CAS): \url{https://www.casact.org/publications-research/research/research-resources/loss-reserving-data-pulled-naic-schedule-p}.

These resources provide the necessary code and data to replicate the study’s findings, in accordance with the Research Transparency policy of the \textit{Annals of Actuarial Science}.

\section*{Competing Interest Statement}
The authors declare no competing interests.

\section*{Funding Statement}
This research did not receive any specific grant from funding agencies in the public, commercial, or not-for-profit sectors.

\bibliographystyle{plain}

\appendix
\renewcommand{\thesection}{\Alph{section}} 

\section{Risk Metric Derivations}
\addcontentsline{toc}{section}{Appendix A: Risk Metric Derivations}
\label{appendix:cvar_derivation}

Conditional Value-at-Risk (CVaR), also known as Expected Shortfall, is a coherent risk measure that captures the average of extreme losses beyond a specified quantile threshold. Unlike the Value-at-Risk (VaR), which only provides a cutoff at a confidence level, CVaR accounts for the full distribution of tail events, making it especially valuable in insurance, finance, and reinforcement learning applications where robustness to rare but catastrophic outcomes is essential \cite{rockafellar2000optimization}.

\subsection*{Value-at-Risk (VaR)}

Let \( L \) denote a real-valued random variable representing a loss. For a given confidence level \( \alpha \in (0,1) \), the Value-at-Risk at level \( \alpha \), denoted \( \text{VaR}_\alpha(L) \), is defined as the smallest threshold \( \ell \) such that the probability of a loss not exceeding \( \ell \) is at least \( \alpha \):

\begin{equation}
\text{VaR}_\alpha(L) = \inf \left\{ \ell \in \mathbb{R} \,\middle|\, \mathbb{P}(L \leq \ell) \geq \alpha \right\}.
\end{equation}

VaR identifies the \(\alpha\)-quantile of the loss distribution, but it fails to account for the severity of losses in the tail beyond this cutoff.

\subsection*{Conditional Value-at-Risk (CVaR)}

CVaR addresses this limitation by measuring the expected loss conditional on exceeding the VaR threshold. The standard definition is:

\begin{equation}
\text{CVaR}_\alpha(L) = \mathbb{E}[L \mid L \geq \text{VaR}_\alpha(L)].
\end{equation}

For continuous loss distributions, CVaR admits an integral representation:

\begin{equation}
\text{CVaR}_\alpha(L) = \frac{1}{1 - \alpha} \int_{\alpha}^{1} \text{VaR}_\tau(L) \, d\tau,
\end{equation}

which reflects the average of quantile values over the distribution’s upper tail beyond the level \( \alpha \).

\subsection*{Optimization-Friendly Reformulation}

Rockafellar and Uryasev \cite{rockafellar2000optimization} proposed an equivalent convex formulation of CVaR, which is particularly well-suited for machine learning and reinforcement learning contexts:

\begin{equation}
\text{CVaR}_\alpha(L) = \min_{z \in \mathbb{R}} \left\{ z + \frac{1}{1 - \alpha} \mathbb{E}[(L - z)^+] \right\},
\end{equation}
where \( (x)^+ = \max(0, x) \) denotes the positive part. The minimizer \( z^* \) corresponds to the VaR at level \( \alpha \), and the overall expression quantifies the expected tail loss. This formulation is convex and differentiable, enabling use with stochastic gradient methods in reinforcement learning pipelines.

\subsection*{CVaR Optimization in Reinforcement Learning}

In a reinforcement learning setting, let \( \tau \sim \pi \) denote a trajectory under policy \( \pi \), and define the cumulative loss as:
\begin{equation}
L(\tau) = \sum_{t=1}^T \ell(s_t, a_t),
\end{equation}
where \( \ell(s_t, a_t) \) is the instantaneous loss (e.g., shortfall or penalty). The CVaR-aware objective becomes:

\begin{equation}
\pi^* = \arg\min_{\pi} \, \text{CVaR}_\alpha(L(\tau)) = \arg\min_{\pi} \, \mathbb{E}[L(\tau) \mid L(\tau) \geq \text{VaR}_\alpha(L(\tau))].
\end{equation}

This formulation enables the agent to prioritize safety and solvency by minimizing expected costs in high-loss scenarios. Gradient-based methods such as actor-critic or policy gradient algorithms can estimate gradients with respect to the CVaR objective using quantile sampling and empirical loss statistics \cite{shapiro2009lectures, sutton2018reinforcement}.

\subsection*{Adaptive CVaR Thresholds in Volatile Regimes}

To enhance tail sensitivity under varying macroeconomic regimes, we introduce a volatility-adaptive CVaR quantile:
\begin{equation}
\alpha_t = 0.90 + 0.05 \cdot \min(1, V_t),
\end{equation}
where \( V_t \in [0,1] \) is a normalized volatility index derived from claim development variance. This dynamic adjustment ensures that the agent becomes more risk-averse in volatile environments by targeting deeper tail percentiles, thus aligning reserve decisions with stress conditions.

\vspace{4mm}
\noindent
{Conclusion.} CVaR provides a principled, tractable, and coherent risk measure that captures worst-case outcomes in a probabilistically sound manner. Its optimization structure supports robust policy learning in risk-sensitive reinforcement learning applications, especially in domains such as insurance reserving, where solvency and capital adequacy under uncertainty are critical.

\section{Algorithm Details}
\addcontentsline{toc}{section}{Appendix B: Algorithm Details}
\label{appendix:algorithm}

This appendix presents the complete training procedure for the proposed regime-aware, CVaR-constrained reinforcement learning framework for insurance reserving. The algorithm corresponds to the PPO-based policy optimization method described in Section~\ref{sec:ppo}.

The agent is trained using Proximal Policy Optimization (PPO) \cite{schulman2017proximal}, with a custom reward function that incorporates penalties for reserve shortfall, capital inefficiency, and violations of regulatory capital thresholds. Conditional Value-at-Risk (CVaR) is estimated using an empirical sampling approach as detailed in Section~\ref{sec:reward_function}.

To promote robustness under stress scenarios, the agent is trained across a macroeconomic curriculum of progressively adverse regimes, indexed by volatility level $\ell \in \{0, 1, 2, 3\}$. This curriculum-based exposure improves generalization across economic conditions and enhances tail-risk sensitivity.

\vspace{1em}
\begin{algorithm}[H]
\begin{algorithmic}[1]
\State {Input:} Claims dataset \( \mathcal{D} \); curriculum levels \( \ell \in \{0, 1, 2, 3\} \); PPO hyperparameters (see Section~\ref{sec:training-hyperparams})
\State {Initialize:} Policy network \( \pi_\theta \), value network \( V_\phi \), violation memory \( \nu_t \gets 0 \), shortfall buffer \( \mathcal{B} \gets \emptyset \)
\For{each seed \( s \in \mathcal{S} \)}
    \For{each curriculum level \( \ell \)}
        \State Sample macro shocks \(M_t \sim \mathcal{N}(\mu_{\ell_t}, \sigma_{\ell_t}^2) \)
        \State Initialize environment \( \mathcal{E}_\ell \) with volatility-adjusted claims
        \For{each PPO rollout episode}
        	    \State Run episode for horizon \(T\) (full development periods) or until terminal development year is reached
            \State Observe state \( s_t = \{ R_t, L_t, V_t, K_t, \nu_t, M_t, \ell_t \}\)
            \State Sample action \( a_t \sim \pi_\theta(s_t) \)
            \State Apply reserve adjustment and update the environment
            \State Compute reward components:
            \begin{align*}
                \hat{S}_t &= \max(0, L_t - R_t) && \text{(shortfall)} \\
                \hat{C}_t &= |R_t - L_t| && \text{(capital inefficiency)} \\
                R_t^{\mathrm{reg}} &= 0.4 + 0.2 \cdot V_t && \text{(solvency floor)} \\
                \alpha_t &= 0.90 + 0.05 \cdot \min(1, V_t) && \text{(adaptive CVaR level)}
            \end{align*}
            \State Add $\hat{S}_t$ to shortfall buffer \( \mathcal{B} \)
            \State Estimate empirical quantile \( \text{VaR}_{\alpha_t} \) from \( \mathcal{B} \)
            \State Compute CVaR:
            \[
            \widehat{\text{CVaR}}_t = \frac{1}{|\mathcal{T}|} \sum_{s \in \mathcal{T}} \hat{S}_s \quad \text{where } \mathcal{T} = \left\{ s : \hat{S}_s \geq \text{VaR}_{\alpha_t} \right\}
            \]
            \State Compute total reward:
            \[
            r_t = -\left( \lambda_1 \hat{S}_t + \lambda_2 \widehat{\text{CVaR}}_t + \lambda_3 \hat{C}_t + \lambda_4 \cdot \mathbb{I}[R_t < R_t^{\mathrm{reg}}] \right)
            \]
            \State Update violation memory:
            \[
            \nu_t \gets 0.95 \cdot \nu_{t-1} + 0.05 \cdot \mathbb{I}[R_t < R_t^{\mathrm{reg}}]
            \]
        \EndFor
        \State Update \( \pi_\theta \), \( V_\phi \) using PPO optimizer
    \EndFor
\EndFor
\State {Return:} Trained policy \( \pi_\theta \) and regime-stratified performance metrics
\end{algorithmic}
\caption{Training Algorithm for RL-CVaR with Regime-Aware Curriculum}
\label{alg:rl_cvar}
\end{algorithm}
\vspace{1em}

This algorithm implements risk-sensitive reserve optimization under dynamic volatility and solvency constraints. The CVaR estimate $\widehat{\text{CVaR}}_t$ reflects the average of the worst-case shortfall outcomes at time $t$, based on empirical quantiles sampled from past rollouts (see Section~\ref{sec:cvar_estimation}). By gradually increasing exposure to more volatile regimes during training, the framework ensures that the learned policy generalizes to macroeconomic stress scenarios consistent with Solvency II and ORSA protocols \cite{eiopa2014solvency, rockafellar2000optimization}.



\section{Glossary of Acronyms}
\addcontentsline{toc}{section}{Appendix C: Glossary of Acronyms}
\label{appendix:glossary}

\begin{table}[!htbp]
\centering
\footnotesize
\begin{tabular}{|p{3cm}|p{9cm}|}
\hline
\textbf{Acronym} & \textbf{Description} \\
\hline
A2C & Advantage Actor-Critic (RL algorithm) \\
AutoRL & Automated Reinforcement Learning \\
BFM & Bornhuetter-Ferguson Model \\
CAS & Casualty Actuarial Society \\
CES & Capital Efficiency Score \\
CLM & Chain-Ladder Model \\
CVaR & Conditional Value-at-Risk \\
DRL & Deep Reinforcement Learning \\
EMA & Exponential Moving Average \\
Gymnasium & Reinforcement learning environment API (formerly OpenAI Gym) \\
IFRS 17 & International Financial Reporting Standard 17 \\
MDP & Markov Decision Process \\
ORSA & Own Risk and Solvency Assessment \\
PPO & Proximal Policy Optimization \\
RAR & Reserve Adequacy Ratio \\
RAROC & Risk-Adjusted Return on Capital \\
RL & Reinforcement Learning \\
RVR & Regulatory Violation Rate \\
VaR & Value-at-Risk \\
\hline
\end{tabular}
\caption{Glossary of Acronyms Used in the Paper}
\label{tab:acronym_glossary}
\end{table}

\end{document}